\documentclass[a4paper,fleqn]{tex/cas-dc}

\usepackage[numbers,sort&compress]{natbib}

\usepackage{lineno}

\def\tsc#1{\csdef{#1}{\textsc{\lowercase{#1}}\xspace}}
\tsc{WGM}
\tsc{QE}

\usepackage[utf8]{inputenc}

\usepackage[colorinlistoftodos]{todonotes} 
\usepackage{subfig}

\usepackage{booktabs}

\usepackage{rotating}
\usepackage{multirow}
\usepackage{siunitx}

\DeclareMathOperator*{\argmax}{argmax}
\DeclareSIUnit\Px{Px}
\DeclareSIUnit\gms{\gram\per\meter\squared}
\DeclareSIUnit\mm{\milli\meter}
\DeclareSIUnit\bit{bit}
\DeclareSIUnit{\kilonewton}{\kilo\newton}
\usepackage{romannum}
\usepackage{graphicx}
\usepackage{lipsum}  

\usepackage{nicefrac}

\usepackage{algorithm}
\usepackage{algpseudocode}
\usepackage{tabularx}
\newcolumntype{Y}{>{\centering\arraybackslash}X}

\newcommand{\R}[1]{\textsuperscript{\textregistered}}

\usepackage{todonotes}
\usepackage{listings}

\usepackage{algorithm}
\usepackage{algpseudocode}

\usepackage{ulem}
\usepackage{float}

\usepackage{tikz}
\usepackage{pgfplots}
\pgfplotsset{compat=1.18}
\usetikzlibrary{arrows,decorations.pathreplacing,decorations.markings,intersections, pgfplots.fillbetween,shapes.geometric,positioning}

\newcommand{\tikzArrow}{%
        \begin{tikzpicture}[inner sep=0pt,baseline=(base)]%
        \draw[-latex,thick](0,0) -- (8mm,0); 
        \node (base) at (0,-.5ex) {};
        \end{tikzpicture}%
    }

\pgfplotsset{
    every axis/.style={
        label style={font=\color{black}},
        tick label style={font=\color{black}},
        title style={font=\color{black}}
    },
 every axis legend/.append style={
    font=\color{black},
    /tikz/every node/.style={text=black}
},   
every node/.style={text=black,/tikz/every node/.style={text=black}}
}

\definecolor{plasma1}{HTML}{0C0786}  
\definecolor{plasma2}{HTML}{350498}  
\definecolor{plasma3}{HTML}{6300A7}  
\definecolor{plasma4}{HTML}{8B0AA5}  
\definecolor{plasma5}{HTML}{B83289}  
\definecolor{plasma6}{HTML}{DB5C68}  
\definecolor{plasma7}{HTML}{F48849}  
\definecolor{plasma8}{HTML}{F7E225}  

\colorlet{type1}{plasma2}
\colorlet{type2}{plasma4}
\colorlet{type3}{plasma7}

\begin{document}
\let\WriteBookmarks\relax
\def\floatpagepagefraction{1}
\def\textpagefraction{.001}

\shorttitle{Machine Learning and Descriptor-Based Analysis of Textile Compaction}

\shortauthors{Düreth et~al.}

\title [mode = title]{Analysis of the Compaction
Behavior of Textile Reinforcements in Low-Resolution In-Situ CT Scans via Machine-Learning and Descriptor-Based Methods}

\author[1]{Christian Düreth}[orcid=0000-0002-6817-1020] 
\cormark[1]
\ead{christian.duereth@tu-dresden.de}
\credit{ Conceptualization, Data curation, Investigation, Methodology, Software, Formal analysis, Validation, Funding acquisition, Visualization, Writing - original draft}

\author[1]{Jan Condé-Wolter}[orcid=0000-0002-1056-5668] 
\credit{Conzeptualization, Funding acquisition, Writing - Review \& Editing}

\author[1]{Marek Danczak}[] 
\credit{Methodology, Investigation, Data Curation, Writing - review \& Editing}

\author[1]{Karsten Tittmann}[orcid=0000-0002-2280-7580]
\credit{Conceptualization, Writing - Review \& Editing}

\author[1]{Jörn Jaschinski}[]
\credit{Supervision, Conceptualization, Investigation, Writing - Review \& Editing}

\author[1,2]{Andreas Hornig}[orcid=0000-0003-2653-7546] 
\credit{Methodology, Writing - Review \& Editing} 

\author[1]{Maik Gude}[orcid=0000-0003-1370-064X]
\credit{Conceptualization, Funding acquisition, Resources, Writing - review \& editing} 

\affiliation[1]{organization={Institute of Lightweight Engineering and Polymer Technology},
    addressline={Dresden University of Technology}, 
    city={Dresden},
    postcode={01307}, 
    country={Germany}}
    
\affiliation[2]{organization={Center for Scalable Data Analytics and Artificial Intelligence Dresden/Leipzig (ScaDS.AI)},
    addressline={Dresden University of Technology}, 
    city={Dresden},
    postcode={01069}, 
    country={Germany}}

\cortext[cor1]{Corresponding author}


\begin{abstract}
A detailed understanding of material structure across multiple scales is essential for predictive modeling of textile-reinforced composites. 
Nesting—characterized by the interlocking of adjacent fabric layers through local interpenetration and misalignment of yarns—plays a critical role in defining mechanical properties such as stiffness, permeability, and damage tolerance. 
This study presents a framework to quantify nesting behavior in dry textile reinforcements under compaction using low-resolution computed tomography (CT). 
In-situ compaction experiments were conducted on various stacking configurations, with CT scans acquired at \SI{20.22}{\micro\meter} per voxel resolution. 
A tailored 3D-UNet enabled semantic segmentation of matrix, weft, and fill phases across compaction stages corresponding to fiber volume contents of \qtyrange[]{50}{60}{\percent}. 
The model achieved a minimum mean Intersection-over-Union of \num{0.822} and an $F1$ score of \num{0.902}. 
Spatial structure was subsequently analyzed using the two-point correlation function $S_2$, allowing for probabilistic extraction of average layer thickness and nesting degree. 
The results show strong agreement with micrograph-based validation. 
This methodology provides a robust approach for extracting key geometrical features from industrially relevant CT data and establishes a foundation for reverse modeling and descriptor-based structural analysis of composite preforms.
\end{abstract}


\begin{highlights}
\item In-situ compaction experiments of dry textile reinforcements with varying layer configurations
\item Semantic segmentation of low-resolution computed tomography scans using a 3D-UNet architecture
\item Descriptor-based spatial analysis for enhanced characterization of material phase distributions
\item Probabilistic analysis of the nesting behavior in multilayer textile reinforcements
\end{highlights}

\begin{keywords}
in-situ computer tomography \sep textile reinforced composites \sep semantic segmentation \sep machine learning \sep descriptor analysis 
\end{keywords}

\maketitle

\section{Introduction}

A detailed analysis of material structures is essential for advancing material science. 
Understanding the relationships between different scales, such as the micro- and meso-level, and their influence on effective material properties is a key aspect of multi-scale computational frameworks~\cite{seibert_statistical_2024,paepegem_1_2021}. 
In composite materials, where constituents with distinct properties are combined, accurately predicting behavior relies on a thorough understanding of their spatial distribution.
The modeling of compaction behavior has also been extensively investigated in recent literature, providing valuable insights into structural evolution during processing~\cite{thompson_high_2018,sun_characterizing_2025}.
Especially, textile reinforced composites where the arrangement of adjacent layers within a composite material, referred to as nesting, is of utmost importance.
Nesting describes how individual plies or fabric layers settle relative to each other during manufacturing~\cite{potluri_compaction_2008}.
The degree of nesting affects key material characteristics such as stiffness, permeability, damage tolerance, and damage~\cite{varandas_importance_2020,dureth_determining_2020,olave_mode_2015}.
Hence, it is of utmost importance to identify, characterize, and analyze the heterogeneous structure of textile reinforced composites. 

Advanced non-destructive testing (NDT) methods like computed tomography (CT) enhance this analysis, but distinguishing individual components from the absorption spectrum remains challenging. 
This challenge is even greater when segmenting individual yarn orientations in textile-reinforced composites, which is essential for any subsequent numerical analysis predicting effective properties. 
However, the usage of very high resolutions (down to a voxel size of \SI{1}{\micro\meter}) can overcome this difficulty by using the structure tensor analysis to rapidly and accurately estimate the local fiber orientations~\cite{auenhammer_robust_2022,straumit_micro-ct-based_2018,jeppesen_characterization_2020}.
Segmentation into individual fiber yarns is done in a post-processing step by clustering  obtained orientations, anisotropies, etc. with advanced techniques like k-means~\cite{straumit_quantification_2015,lomov_meso-fe_2007}.
It is reported in~\cite{ali_efficient_2022} that the analysis of structure tensor analysis is limited to resolution $<$\SI{5}{\micro\meter}.
Noteably, high-resolution scans such as $\mu$CT are expensive and restricted to small volumes, making it difficult to capture sufficient data across multiple textile repeating units.

Recent advancements in machine learning and deep learning methods have leveraged the advanced analysis methods in material science.
The range of methods includes segmentation techniques for detecting cracks in composites from CT scans~\cite{helwing_deep_2022}, cell structure analysis and density-dependent bead evaluation~\cite{koch_analysis_2023}, efficient segmentation methods for textile reinforced composites in weft and fill of $\mu$CT scans~\cite{ali_efficient_2022}, and encoding approaches using mixed convolutional and recurrent neural networks to predict effective properties~\cite{koptelov_deep_2024}. 
Additionally, generative techniques such as super-resolution techniques for enhancing CT scans~\cite{guo_deep-learning_2023}, removing CT artifacts by leveraging the capabilities of generative artificial networks by an inpainting method~\cite{karamov_inpainting_2021}, and diffusion-based reconstruction methods~\cite{dureth_conditional_2023,lyu_microstructure_2024} further expand the capabilities of machine learning methods for advanced material characterization.

For segmentation tasks, the usage of convolutional neural networks (CNNs) is most popular in recent literature as they are highly effective in capturing spatial features and patterns within material structures~\cite{bishop_deep_2023}.
Their ability to learn hierarchical representations from image or volume data leverages modern segmentation methods~\cite{badran_automated_2020}.
The UNet is a widely used CNN architecture designed for biomedical image segmentation but has since been adapted for various applications, including material science and composite analysis. 
Originally introduced by Ronneberger et al.~\cite{ronneberger_u-net_2015}, UNet follows an encoder-decoder structure that enables precise pixel-wise segmentation in binary and even multi classification tasks while retaining spatial context.
Extensions of UNet have been developed to improve segmentation performance in different domains. 
For instance, 3D UNet~\cite{cciccek_3d_2016} extends the original architecture to volumetric data, making it well-suited for segmenting CT scans from textile reinforced composite. 
ResUNet (rUNet)~\cite{jha_resunet_2019} integrates residual connections to improve gradient flow and enhance training stability, especially for deeper networks~\cite{bishop_deep_2023}.


After successfully segmenting the target constituents, the next step is to extract their geometric characteristics, which serve as the foundation for quantitative analysis. 
Traditional 1-dimensional (1-D) descriptors such as volume fraction, surface area, and aspect ratio provide essential insights.  
For textile reinforcements, 1-D descriptors such as crimp angle, nesting factor, as introduced by Potluri~\cite{potluri_compaction_2008}, cross-section properties like major and minor diameter, spacing between adjacent yarns, etc. have been commonly used to characterize and model layer arrangements and interlocking behavior~\cite{long_modelling_2011,liu_multi-scale_2016,yousaf_compaction_2021}.
While these descriptors effectively capture basic geometric properties, they often fail to  describe the complex spatial relationships within textile composites fully~\cite{torquato_random_2002}.

Higher-order or $n$-D descriptors enable a more detailed and probabilistic understanding of material structure and its influence on mechanical behavior as shown in \cite{seibert_statistical_2024}. 
Python-based open-source libraries like PyMKS~\cite{brough_materials_2017} and MCRpy~\cite{seibert_microstructure_2022-1} provide higher-order descriptor-based analysis tools.
As PyMKS features only fast FFT-based implementation of the two-point (auto)-correlation and tools like principal-component-analysis (PCA) for structure-property linkages, MCRpy offers a variety of descriptors and gradient-based optimization algorithms for reconstruction~\cite{seibert_reconstructing_2021,dureth_conditional_2023}.
Beneficial to $n$-D descriptor like the FFT-based 2-point correlation is that they are directly applicable to the segmented voxel mesh, which reduces the necessity for post-processing of the segmentation map.

In this study, we perform in-situ CT scans at different load levels to analyze the compaction behavior of a carbon-fiber plain-weave textile manufactured from Tenax\textsuperscript{\textregistered}~–~E HTA40 E13 yarn.
The considered areal weight are \SI{245}{\gms}, which leads to higher ondulation, less nesting capabilities as shown in~\cite{dureth_determining_2020}.
Thereby, we achieved with the test setup a spatial resolution of \SI[]{20.22}{\micro\meter\per voxel}. 
This resolution is approximately 3 times higher than the average fiber diameter of \SI{7}{\micro\meter}, making it difficult for segmentation via structure tensor analysis.
Hence, a modified 3D-UNet was trained with customized loss functions to classify the obtained CT scans in weft, fill, and background using the multi-GPU PyTorch implementation from~\cite{cciccek_3d_2016,wolny_accurate_2020}. 
Our fork from the repository additionally implemented a fast FFT-based 2-point (auto-)~correlation, useful data import and export functionalities like \texttt{*.nrrd} and \texttt{*.vti} for visualization, and tensorboard integration for tracking of training metrics. 
Finally, all segmented textile material's label maps are characterized using descriptors and compared to classical evaluation from~\cite{dureth_determining_2020}, giving a more detailed insight into the compaction behavior of textile-reinforced composites from low-resolution in-situ compaction experiments with emphasis on the nesting behavior.


\section{Experimental Methods}
\label{sec:experimental_methods}

\subsection{Materials}
\label{ssec:materials}

\begin{figure}
    \centering
    \includegraphics[page=1,width=.4\textwidth]{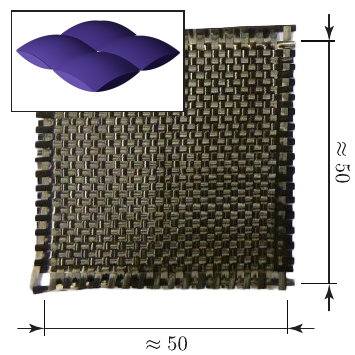}
    \caption{Photograph of a plain-weave carbon fiber fabric sample and its corresponding idealized unit cell geometry modeled in TexGen~\cite{long_modelling_2011} (dimensions are in \si{\milli\meter}).}
    \label{fig:material}
\end{figure}

In this study, a single plain-weave textile material was selected, representative of the type investigated in~\cite{dureth_determining_2020}. The textile is produced from Tenax\textsuperscript{\textregistered}~–~E HTA40 E13 yarn, composed of \num{3000} continuous carbon filaments with a linear density of \SI{200}{tex}. 
Each filament has a diameter of approximately \SI{7}{\micro\meter}. 
The yarn consists of high-strength, standard-modulus carbon fibers developed for aerospace and industrial applications, exhibiting a tensile strength of \SI{4100}{\mega\pascal} in fiber direction, a maximum elongation of \SI{1.7}{\percent}, and a Young’s modulus of \SI{240}{\giga\pascal}.
A thread density of \SI{7}{threads\per\centi\meter} results in an areal weight $\rho_{A}$ of \SI{285}{\gram\per\metre\squared} and an approximate layer thickness of \SI{0.38}{\milli\meter}, determined in accordance with DIN EN 12127 and DIN EN ISO 5084, respectively. 
The material has a fiber density $\varrho_f$ of \SI{1.77}{\gram\per\cubic\centi\meter}. Figure~\ref{fig:material} illustrates a single-layer fabric specimen with a dimension of \qtyproduct{50x50}{\milli\meter}, along with an idealized geometric representation of a unit cell, generated using TexGen~\cite{long_modelling_2011}.

All textile layers were cut into a quadratic shape of \qtyproduct{50x50}{\milli\meter} using a CNC cutting table with an oscillating diamond-coated blade for precision. 
The stacking of \qtylist[]{1;5;10;37}{} layers for each experiment was manually performed with optical alignment to ensure accurate positioning.

\subsection{In-situ CT Compactation Experiment}
\label{ssec:insitu-CT}

All compaction experiments were performed using an in-situ computed tomography (CT) testing rig as described in~\cite{bohm_quantitative_2015}. 
This custom-built testing system features a Zwick Z250 universal testing machine (UTM) and a Finetec CT FCTS 160, enabling high-precision mechanical testing under real-time X-ray imaging.
The Zwick Z250 UTM supports both tensile and compressive loads up to \SI{250}{\kilo\newton}, as well as torsional loads up to \SI[inter-unit-product = \mkern-1mu]{2}{\kilonewton\meter}, facilitating a comprehensive range of mechanical testing conditions. 
The Finetec CT FCTS 160 features a \SI{160}{\kilo\volt} microfocus X-ray tube with a minimum focal spot size below \SI{3}{\micro\meter}, coupled with a high-resolution flat panel detector. 
The detector has a resolution of \qtyproduct{3200x2300}{\Px} with \SI{16}{\bit} depth, where each pixel measures \qtyproduct{127x127}{\micro\meter}.

\begin{figure*}[tb]
    \centering
    \includegraphics[page=1,width=.9\linewidth]{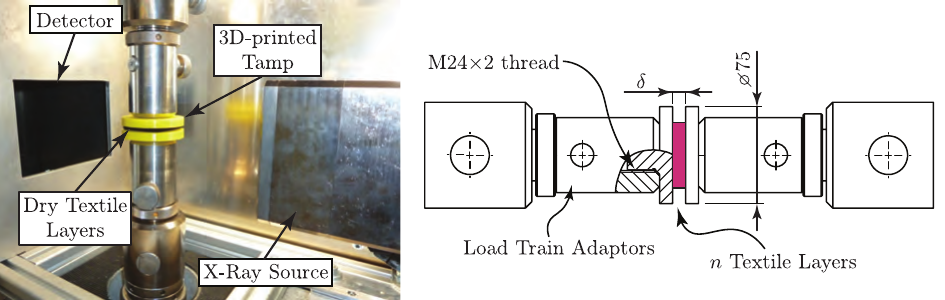}
    \caption{Experimental set-up (left) and drawings of the load train (right) in situ CT test set-up (all dimensions are in \si{\milli\meter})}
    \label{fig:exp-setup}
\end{figure*}

A distinguishing feature of this setup is the use of a cone-beam CT arrangement, where both the X-ray source and detector are mounted on a precision-engineered marble rotation table instead of rotating the specimen (cf. Figure~\ref{fig:exp-setup} (left)). 
This configuration allows for flexible imaging conditions but also leads to variations in resolution, magnification, and the imaged specimen volume, also referred to as the region of interest (ROI), depending on the source-to-object and source-to-detector distances.
By adjusting these parameters, different levels of detail can be achieved, optimizing the trade-off between field of view and spatial resolution for specific experimental requirements.

\begin{table}[h]
    \centering
    \caption{Overview of tamp distance $\delta_n$ for each load stage and layer configuration $n$ with respect to fiber volume fraction $\phi$ ($\rho_A=\SI{285}{\gms}$; $\varrho_f=\SI{1.77}{\gram\per\cubic\centi\meter}$).}
    \label{tab:test_matrix}
    \begin{tabular}{c|c c c c} 
    \toprule
         $\phi$ & $\delta_{1}$ & $\delta_{5}$ & $\delta_{10}$ & $\delta_{37}$\\
         $[-]$  & $[\si{\mm}]$ & $[\si{\mm}]$ & $[\si{\mm}]$  & $[\si{\mm}]$ \\
         \midrule
        $<$ \num{.30} & \num{1.87} & \num{4.49} & \num{6.84} & \num{20.12} \\
         \num{.50}& \num{0.32} & \num{1.62} & \num{3.24} & \num{11.98} \\
         \num{.55}& \num{0.29} & \num{1.47} & \num{2.94} & \num{10.89} \\
         \num{.60}& \num{0.27} & \num{1.35} & \num{2.70} & \num{10.00} \\
         \bottomrule
    \end{tabular}
\end{table}

As previously outlined, this study investigates textile layers shaped as \qtyproduct{50x50}{\milli\meter}, which are stacked to form specimens composed of \qtylist{1;5;10;37}{} layers. 
To accommodate these geometries during mechanical testing, a custom load train setup was designed, as illustrated in Figure~\ref{fig:exp-setup}.
The compression tamps, each with a diameter of \SI{70}{\milli\meter}, were fabricated using a high-resolution desktop 3D printer. Polylactide (PLA) was selected as the printing material to ensure sufficient mechanical performance under compressive loads. Despite its relatively high density of \SI{1.24}{\gram\per\cubic\centi\meter}, PLA offers a favorable balance of stiffness and strength, with a Young’s modulus of \SI{3300}{\mega\pascal} and a tensile strength of \SI{110}{\mega\pascal}, making it well-suited for this application.
For precise alignment within the load train, each tamp was mounted via an M24$\times$2 thread, which was machined directly onto the tamp structure, as shown in Figure~\ref{fig:exp-setup}. 
This setup ensured both mechanical integrity and reproducibility across all tested specimen configurations and for a load limit of \SI{2.5}{\kilo\newton}. 
Lastly, according to the test set-up, this leads to the resolutions of \SI[]{20.22}{\micro\meter\per\Px} and magnifications of \num[]{10.9}.

For comparability of the load stages, we aimed to achieve equivalent fiber volume contents $\phi$ (cf. Eq.~\ref{eq:volume_content}) for each textile layer configuration. 
The thickness $\delta$ of a laminate, as a function of $\phi$, given a specified number of layers $n$, the fiber density $\varrho_f$, and the areal density $\rho_A$, is expressed as
\begin{align}
\delta_n(\phi) = \frac{n \ \rho_{A}}{\varrho_{f}} \ \frac{1}{\phi}.
\end{align} 
The parameter $\delta$ represents the gap between the tamp and the load stages as depicted in Figure~\ref{fig:exp-setup}, which is critical for CT scans and serves as a displacement control parameter for the testing machine. 
Further on to compensate for any setting of the textile layers, a waiting time of \SI{8}{\min}  was incorporated before each scan in the testing procedure.
For each layer configuration four load stages of $\phi =$ \qtylist[]{45;50;55;60}{\percent} have been identified to be suitable for this study.
Table~\ref{tab:test_matrix} lists all calculated $\delta$ for each load step and layer configuration.
Maintaining a consistent gap and no displacement of the fiber roving ensures precise imaging, comparability of different load stages, and accurate regulation of the compaction process, allowing for controlled fiber volume content and reproducibility across different material configurations.

Each CT scan was conducted using 1440 acquisition steps, capturing projections over a full rotation from \qtyrange[]{0}{360}{\degree} on the fly.
The acquisition time for each projection was \SI{625}{\milli\second}, ensuring high-quality imaging with minimal motion artifacts.
The X-ray source operated at \SI{140}{\kilo\volt} and \SI{80}{\micro\ampere}, ensuring sufficient X-ray attenuation contrast for accurate material differentiation and structural resolution.
The acquired projection data were subsequently reconstructed using CERA-BASIC, which utilizes filtered back projection (FBP) as its core algorithm. 
This method enables rapid and robust reconstruction while incorporating essential correction techniques such as beam hardening compensation, grid-based tomography, and geometric calibration via projection matrices.
The detector settings and exposure parameters were optimized to achieve a balance between signal quality and noise reduction, allowing for precise visualization of textile layers.

\section{Numerical Methods}
\label{sec:numerical_methods}

\subsection{3D-UNet}
\label{ssec:model_unet}

\begin{figure*}[b]
    \centering
    \includegraphics[width=\linewidth]{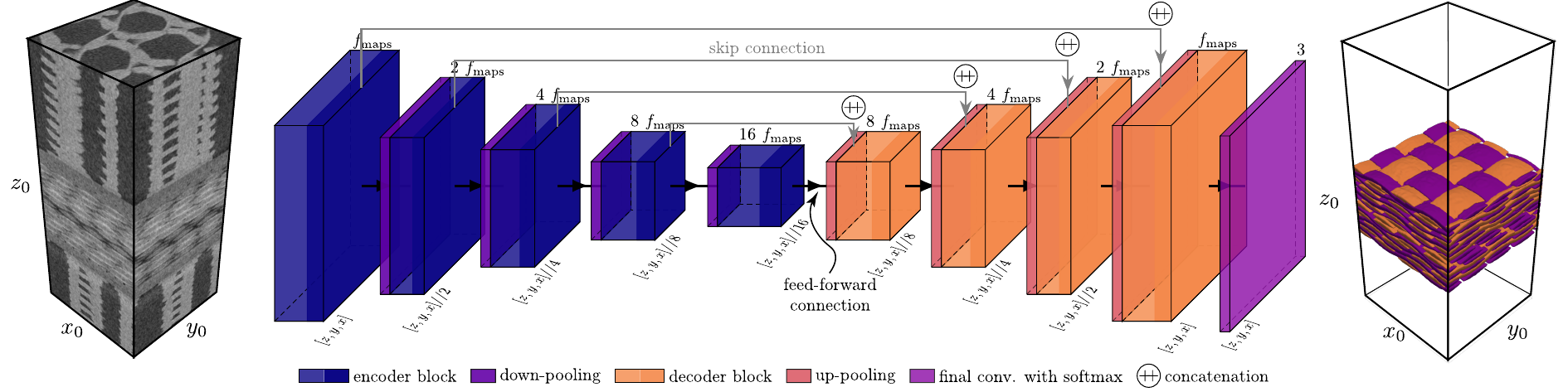}
    \caption{Schematic illustration of 3D-UNet's architecture for semantic segmentation of textile reinforcement as a three-class semantic segmentation problem (Illustration was prepared using code from~\cite{iqbal_harisiqbal88plotneuralnet_2018})}
    \label{fig:model_runet}
\end{figure*}

The CNN model of a 3D-UNet consists of an encoder and decoder. 
It is a common model for advanced segmentation tasks~\cite{wolny_accurate_2020,cciccek_3d_2016,diakogiannis_resunet-_2020} and for generative applications such as stable diffusion as shown in~\cite{dureth_conditional_2023,bishop_deep_2023}.
Therefore, the encoder and decoder architecture consists of so-called building blocks for encoding and decoding. 
Figure~\ref{fig:model_runet} schematically depicts the architecture of a 3D-UNet, highlighting the spatial dimensions $[x, y, z]$ and the depth of feature maps $f_{\text{maps}}$, expressed as multiples of the initial encoder block for each building block.

The fundamental concept of a UNet is to learn which features are essential for effective encoding and decoding of spatial information~\cite{bishop_deep_2023}. 
This is achieved through a symmetric encoder–decoder structure, where the encoder progressively compresses spatial information to extract high-level features, while the decoder reconstructs the output using a series of upsampling and convolutional operations. 
Crucially, skip connections between corresponding encoder and decoder levels enable the network to preserve fine-grained spatial details, improving segmentation accuracy, particularly in scenarios with limited training data, high spatial complexity, or noise.

Each encoder block consists of a down-pooling layer, followed by two times the sequence of a group normalization layer (\texttt{groupnorm3D}\footnote{\texttt{groupnorm3D} normalizes over groups rather than over batches advantageous in segmentation task with small batch sizes via: $\hat{x} = (x - \mu_G)/(\sqrt{\sigma_G^2 + \epsilon})$ with $mu_G$ and $\sigma_G$ the mean and std. deviation over a group the size $[lkm]$ and $\epsilon$ a constant factor for numerical stability}), a ($3\times3\times3$)-convolution (\texttt{conv3D}),  and a rectified linear unit (ReLU) as activation function~\cite{cciccek_3d_2016}. 
This sequence is also called double-convolution.
However, alternative activation functions are also commonly explored in the literature~\cite{bishop_deep_2023,clevert_fast_2016,he_delving_2015,hendrycks_gaussian_2023}. 
For instance, Leaky ReLU and Parametric ReLU (PReLU) have been shown to improve gradient flow in deeper networks, while ELU (Exponential Linear Unit) and GELU (Gaussian Error Linear Unit) offer smoother non-linearities that may enhance learning dynamics and convergence properties. 
The choice of activation function can significantly impact model performance and is often empirically selected based on the specific characteristics of the data and the learning task. 
Here, we decided on a common ReLU architecture due to its simplicity, computational efficiency, and well-documented effectiveness in segmentation networks~\cite{cciccek_3d_2016,wolny_accurate_2020}.
For spatial down-pooling, we choose a convolutional layer with a stride of 2 rather than a max-pooling layer as proposed in~\cite{wolny_accurate_2020}.
Down-pooling with a convolutional stride is common in model architectures such as ResNets~\cite{he_deep_2015}.
In contrast to max-pooling, it adds more learnable parameters, but it can leverage the down-pooling step as an opportunity to learn robust feature representations, particularly in the presence of noise or subtle spatial variations, and guarantees a better gradient flow in the backpropagation.

The decoder path mirrors the encoder structure in reverse, following a similar sequence of operations. 
At the beginning of each decoder block, the spatial resolution is increased via trilinear interpolation, effectively restoring the original input dimensions in a stepwise manner. 
This upsampling is followed by concatenation with the corresponding feature maps from the encoder path through skip connections.
Concatenation refers to the numerical operation of joining two or more tensors $A\in\mathbb{R}^{C_1\times(spatial)}$ and $B\in\mathbb{R}^{C_2\times(spatial)}$ along side a specific dimension to obtain $\texttt{concat}(A,B)\in\mathbb{R}^{(C_1+C_2)\times(spatial)}$, without performing any arithmetic calculation.
In Figure~\ref{fig:model_runet} the numerical operation concatenation is depicted as the following operator $\mathbin{+\!\!+}$.
Each decoder block then performs the double-convolution operation — defined as two successive sequences of \texttt{groupnorm3D}–\texttt{conv3D}–\texttt{ReLU}, as mentioned earlier — while reducing the number of feature maps $f_{\text{maps}}$ to gradually compress the representation.
The decoder continues in this fashion until the desired number of output feature maps ($f_{maps}=C$) and spatial resolution ($[z,y,x]$) are reached, matching the dimensions required for final classification into the target number of segmentation classes $C$.
Whilst training, the model outputs raw logits, representing unnormalized scores for each class at every voxel. 
In evaluation mode, these logits are converted into class-wise probabilities using the softmax function, enabling voxel-wise predictions. 
The softmax function is defined as:
\begin{align}
\text{softmax}(x)_i=\sigma(x)_i = \frac{e^{x_i}}{\sum_{j=1}^{C} e^{x_j}} \quad \forall\ C > 2\label{eq:softmax}
\end{align}
where $C$ denotes the total number of classes and channels, respectively, and $x_i$ is the logit corresponding to class $i$ at voxel $x$~\cite{bishop_deep_2023}.
The final conversion from class probabilities to a discrete label map is performed by selecting the class with the highest predicted probability at each voxel. This is achieved using the $\argmax$ function:
\begin{align}
\hat{y} = \argmax_c \ p(c)
\end{align}
here $p(c)$ denotes the predicted probability for class $c$, and $\hat{y}$ is the resulting label assigned to each voxel.
This operation reduces the output tensor’s dimensionality from $[C, z, y, x]$ to $[z, y, x]$, where each voxel holds the integer class index $c \in \{0, \ldots, C-1\}$~\cite{bishop_deep_2023}. 
Here, the classes matrix/background, weft, and fill correspond to the class integers $0$, $1$, and $2$, respectively.

The final model architecture with the explicit shapes $[b,f_{maps},z,y,x]$ is listed in Table~\ref{tab:model_parameters}. Here, the output shapes and number of learnable parameters of every encoder and decoder block are listed for an arbitrary batch size $b$ denoted as $-1$.
The codebase is available online.\footnote{\url{https://github.com/choROPeNt/3dseg}}

\subsection{Data}
\label{ssec:data}

The data annotation for both the training and evaluation datasets was conducted using the open-source software 3D Slicer~\cite{fedorov_3d_2012}. 
In this context, three semantic classes were defined to represent the textile reinforcement structure: background/matrix, weft, and fill yarns. 
These segmentations serve as a foundation for subsequent descriptor-based post-processing and analysis.

For practical reasons, only subvolumes of the raw CT data with dimensions $(h\ \times\ 256\ \times\ 256 )\ \text{Px}^3$—where the height $h$ varied depending on the textile architecture and the number of stacked reinforcement layers—were annotated and utilized for training and evaluation purposes. Such a subset volume is exemplarily depicted in the Figure~\ref{fig:model_runet} with its dimensions $z_0$, $y_0$, and $x_0$ 
This selective labeling strategy ensured high-quality segmentations while maintaining annotation efficiency.
Lastly, model-assisted labeling was utilized to provide pre-labels for the work-intensive labeling process.
This method demanded just fine-tuning rather than labeling from scratch.

However, to leverage the diversity of the dataset for training, pixel-level data augmentation methods such as Gaussian blur, Poisson noise, and contrast adjustment were employed. 
These techniques introduce local intensity variations that help the model generalize better to subtle grayscale differences and imaging artifacts commonly found in CT data.
Spatial-level data augmentation, on the other hand, was deliberately avoided. 
Although common in many segmentation tasks, such augmentations (e.g., random rotations, elastic deformations, or affine transformations) are not straightforward in this context. 
This is due to the inherent anisotropic structure of the textile reinforcements: the weft and fill yarns follow well-defined and orthogonal axes. 
Applying arbitrary spatial transformations would distort these directional features, leading to semantic inconsistencies between the input data and corresponding labels. 
In turn, this could result in misleading gradients during training and ultimately degrade the segmentation performance.
Therefore, spatial augmentations were excluded to preserve the structural integrity of the labeled yarn directions, ensuring that the network learns meaningful features aligned with the true physical orientation of the textile architecture.

The data used for training and evaluation is available in \cite{dureth_torch3dseg_2025} and provided in the \texttt{*.h5} file format, which is based on the Hierarchical Data Format (HDF5) and allows efficient storage, compression, and access of large-scale volumetric and metadata-rich datasets in an open-source framework.
In that file container, the following data can be found: 
\begin{itemize}
\item \texttt{volume}: Contains the raw X-ray computed tomography (CT) data, stored as a three-dimensional array of \texttt{uint16} values with shape $[x_0, y_0, z_0]$. This dataset represents the greyscale attenuation values in the scanned volume.
 \item \texttt{labels}: Holds the corresponding ground truth segmentation map, also as a three-dimensional array of \texttt{uint16} values with identical shape $[x_0, y_0, z_0]$. Each voxel is assigned a discrete label index representing the semantic class (e.g., background, weft, or fill yarn).
\item \texttt{masks}: Provides binary masks as a one-hot channel for each semantic class, stored as a four-dimensional boolean array with shape $[c, x_0, y_0, z_0]$, where $c$ denotes the number of distinct classes. This dataset was used for training and evaluation.
\item \texttt{instances}: Provides an instance label map for each yarn in weft and fill direction as a three-dimensional array $[x_0, y_0, z_0]$ of \texttt{uint16}.
\end{itemize}

\subsection{Training and Evaluation}
\label{ssec:training}

To train the 3D-UNet model with $3$ classes, a composite loss function is employed that combines cross-entropy and Dice loss in a linear weighted manner. 
This formulation leverages the strengths of both losses: cross-entropy $\mathcal{L}_{\text{CE}}$ encourages accurate voxel-wise classification, while Dice loss $\mathcal{L}_{\text{Dice}}$ directly optimizes for overlap between predicted and ground truth regions~\cite{bishop_deep_2023,cardoso_generalised_2017}. 
The total loss $\mathcal{L}$ therefore accumulates as:
\begin{align}
    \mathcal{L} = \alpha\ \mathcal{L}_{\text{CE}} + \beta\ \mathcal{L}_{\text{Dice}},\label{eq:loss}
\end{align}
where $\alpha$ and $\beta$ denote the respective weights of the individual loss components. 
In this work, we set $\alpha = 0.3$ and $\beta = 0.7$, placing greater emphasis on the Dice loss to encourage accurate learning of class shapes and spatial structure.

The cross-entropy loss for a multi-class classification problem is computed as the mean over all N voxels and is defined as:
\begin{equation}
\mathcal{L}_{\text{CE}} = \frac{1}{N} \sum_{n=1}^{N} \left\{ - \sum_{c=1}^{C} w_c \log\big( \sigma(x_n)_c \big) \cdot y_{n,c} \right\},
\end{equation}
where C is the number of classes, $x_n \in \mathbb{R}^C$ is the vector of raw logits at voxel $n$, and $\sigma(x_n)_c$ denotes the softmax probability for class $c$ at that voxel according to Equation~\ref{eq:softmax}.
The loss is averaged over all $C$ classes.
The ground truth label $y_{n,c} \in \{0, 1\}$ is one-hot encoded, indicating whether voxel $n$ belongs to class $c$, and $w_c$ is an optional weighting factor associated with every individual class.
Class weights $w_c$ are commonly used to compensate for class imbalance in segmentation tasks. 
In this work, we manually set the weights vector to $w_c = [0.1,\ 0.45,\ 0.45]$ to place greater emphasis on learning the weft and fill classes, while down-weighting the background/matrix class.

The second loss mentioned is the  Dice loss $\mathcal{L}_{Dice}$~\cite{cardoso_generalised_2017}. The loss function for multi-class is defined as follows:
\begin{align}
    &\mathcal{L}_{\text{Dice}} = 1 - \frac{1}{C} \sum_{c=1}^{C} w_c \cdot 
    \frac{2 \sum_{n=1}^{N} p_{n,c} y_{n,c}}{\sum_{n=1}^{N} p_{n,c} + \sum_{n=1}^{N} y_{n,c} + \epsilon} \label{eq:loss_dice}
\end{align}
Here, $p_{n,c} \in [0, 1]$ denotes the softmax probability for class $c$ at voxel $n$ according to Equation~\ref{eq:softmax}, and $y_{n,c} \in \{0,1\}$ is the corresponding one-hot encoded ground truth label. 
The numerator represents the intersection between prediction and ground truth, while the denominator accounts for their combined sizes, ensuring scale-invariant comparison. 
The small constant $\epsilon$ is added for numerical stability. 
Similar to the cross-entropy loss, class weights $w_c = [0.1,\ 0.45,\ 0.45]$ are used to control the contribution of each class.

As a reasonable evaluation metric, we considered the $F1$ score and the mean IoU averaged for each class $c\ \in \ \{0,\dots,C-1 \}$.
While both assess overlap between predicted and ground truth masks, the $F1$ score (Dice coefficient, which corresponds to $1-\mathcal{L}_{\text{Dice}}$ according to Equation~\ref{eq:loss_dice}) emphasizes the balance between precision and recall, making it more sensitive to class imbalance. 
In contrast, the mean IoU provides a stricter assessment by penalizing both false positives and false negatives equally, and is widely adopted in semantic segmentation benchmarks. 
For evaluation, we focused only on the weft and fill classes, treating the matrix as background and excluding it from metric computation. 
Since the weft and fill classes are relatively balanced in our dataset, both metrics provide a fair and consistent evaluation of segmentation performance.

Training and evaluation were conducted on the High-Performance Computing (HPC) infrastructure hosted by ScaDS.AI\footnote{ScaDS.AI: Center for Scalable Data Analytics and Artificial Intelligence Dresden/Leipzig}, located at TU Dresden. The HPC system is equipped with nodes featuring 8$\times$ NVIDIA A100-SXM4 Graphics Processing Units (GPUs), each with \SI{40}{\giga\byte} of VRAM, supported by two AMD EPYC processors with 48 cores each and \SI{1}{\tera\byte} of system memory.
This hardware setup provided sufficient computational resources for efficient training and inference on large volumetric spatial patches of size \qtyproduct{128x128x128}{\Px}, extracted from the full input volumes of size $[z_0, y_0, x_0] = [h, 256, 256]$, cf. Section~\ref{ssec:data}. 
Within the patch extraction, a sliding window with a stride of 48 voxels in each spatial direction was used, yielding a total of 144 overlapping patches per volume of size \qtyproduct{500x256x256}{\Px}.
The selected patch configuration, in combination with the current model architecture, results in a memory footprint of approximately \SI{6}{\giga\byte} during the forward and backward passes with gradient computation. 
This allows for a maximum batch size of six patches during training under the given hardware constraints.
More importantly, the chosen patch size of \qtyproduct{128x128x128}{\Px} and \qtyproduct{2.56x2.56x2.56}{\milli\meter}, respectively,  ensures that each sample contains more than one full undulation of the woven structure, enabling the model to learn spatial patterns in a broader context and reducing the influence of noise, low resolution, and local artifacts. 
This is particularly important in cases where the weft and fill yarns are closely interwoven or in contact at higher FVC, as they may appear as a single homogeneous grayscale region in lower-resolution representations. 
By covering a larger spatial extent, the model is better equipped to disentangle these structurally distinct classes based on their broader geometric context, rather than relying solely on local intensity differences~\cite{cciccek_3d_2016}.

The adaptive momentum weighted (ADAMW) optimizer was employed with an initial learning rate of \num{1e-2} and a weight decay of \num{1e-5} for training~\cite{kingma_adam_2017}. 
A manual seed of 42 was chosen for the training experiments' reproducibility.
To enable progressive refinement of the learning process, a learning rate scheduler was applied. 
This strategy ensured effective fine-tuning as training progressed. 
The network was trained for a total of 50 epochs using a batch size of 6 per GPU.

\subsection{Descriptor-Based Analysis}
\label{ssec:descriptor}

The segmented textile reinforcement can be interpreted as an indicator function in the discretized 3D pixel space $ N =  I \times K \times L$ 
\begin{align}
\mathbf{M} \in \mathcal{M} \varsubsetneq \mathbb{R}^{I \times K \times L}
\end{align}
where $\mathbf{M}$ denotes a binary or multi-class mask representing the predicted material class at each voxel, and $\mathcal{M}$ is the space of admissible segmentations~\cite{seibert_statistical_2024,torquato_random_2002}
Each element of $\mathbf{M}$ assigns a discrete label $c \in \{0, \dots, C-1\}$ to a voxel in $N$, with $c = 0$ representing the background (matrix) and $c > 0$ representing the material phases of interest, such as weft and fill.
For many downstream tasks, particularly statistical microstructure characterization, the mask $\mathbf{M}$ is most effectively represented in a one-hot encoded format, where each class is mapped to a separate binary channel. 
This representation facilitates the application of descriptor functions by ensuring class separability and avoiding artificial correlations introduced by ordinal label encodings.
To quantify the morphology of a microstructure in a statistical and translation-invariant manner, a characterization function
\begin{align}
    f : \mathcal{M} \rightarrow \{D_i\}_{i=1}^{n_d}
\end{align}
is introduced. 
This function maps a microstructure realization $\mathbf{M}$ to a set of $n_d$ different descriptors $D_i$. 
A descriptor is an invariant-based feature representation of the microstructure that captures essential geometric or topological characteristics while abstracting away the specific spatial arrangement of local features. 
These descriptors are constructed to remain stable under translation, rotation, and minor perturbations, thereby enabling robust comparison between different realizations of statistically equivalent structures~\cite{seibert_reconstructing_2021}.

Such representations may range from simple one\-di\-men\-sion\-al descriptors, such as volume fractions or nesting factors ($1\text{-}D$ descriptors), to higher-dimensional ($n\text{-}D$) statistical measures, including two-point correlation functions, orientation distributions, or topological characteristics—depending on the material system and the properties under investigation. 
By embedding the high-dimensional spatial information of $\mathbf{M}$ into a lower-dimensional but informative descriptor space, this framework enables the development of structure–property relationships that are both physically meaningful and generalizable across samples, as shown in~\cite{seibert_microstructure_2022-1,seibert_statistical_2024,schneider_fftbased_2025}

\subsubsection{$1\text{-}D$ Descriptors}
\label{sssec:1d_descriptors}

For a binary label map or one-hot encoded channel, respectively, the volume content $\Phi$ of each class $c$ can be computed as follows:
\begin{align}
    \Phi_c = \frac{1}{N} \ \sum_{i=1}^N \mathbf{M_c(i)} \label{eq:volume_content}
\end{align}
where $N$ is the total number of voxels in the domain, and $\mathbf{M}_c(i) \in \{0,1\}$ indicates whether voxel $i$ belongs to class $c$.
The sum overall $\Phi_c$ is equal to 1~\cite{torquato_random_2002}.

As previously mentioned, the nesting factor according to Polturi~\cite{potluri_compaction_2008} is defined as
\begin{align}
\mathcal{N}_F = \frac{t_s}{\sum_{i=1}^z t_i}\label{eq:nesting}
\end{align}
where $t_s$ denotes the total stack thickness corresponding to the compressed state $\delta$ (cf. Figure~\ref{fig:exp-setup}), and $t_i$ represents the individual thickness of layer $i$.
A nesting factor $\mathcal{N}_F = 1$ indicates an idealized configuration in which all layers are perfectly aligned and uniformly compacted. 
Lower values reflect increased geometric interlocking or nesting between layers.

In \cite{dureth_determining_2020}, a nesting factor of $ \mathcal{N}_F = $ \num[separate-uncertainty=true]{0.868(37)} 
was reported for the textile reinforcement material used in this work, based on micrograph analysis of a consolidated specimen with a FVC of approximately \SI{60}{\percent}.

\subsubsection{$n\text{-}D$ Descriptors}

As previously mentioned, higher-dimensional descriptors $\mathcal{D}^n$, such as the two-point correlation function $S_2$, help capture the probabilistic nature of the spatial arrangement of textile layers. 
The two-point correlation is defined as:
\begin{align}
S_2(\mathbf{r}) = \langle f(\mathbf{x}), f(\mathbf{x} + \mathbf{r}) \rangle
\end{align}
where $f(\mathbf{x})$ is the indicator function of a given phase, and the angular brackets denote spatial averaging~\cite{seibert_microstructure_2022-1}. 
This descriptor quantifies the likelihood of finding two points, separated by vector $\mathbf{r}$, in the same phase, thereby encoding information about phase distribution and spatial correlations.

A practical and efficient implementation of the two-point correlation $S_2$ in numerical settings is based on the autocorrelation of the indicator function $\mathbf{M}$. 
Leveraging the convolution theorem, the computation can be performed in Fourier space as~\cite{seibert_statistical_2024}:
\begin{align}
S_2 = \frac{1}{N} \mathcal{F}^{-1} \left( \mathcal{F}(\mathbf{M}) \odot \mathcal{F}^*(\mathbf{M}) \right) \label{eq:d_s2}
\end{align}
Here, $\mathcal{F}$ and $\mathcal{F}^{-1}$ denote the forward and inverse discrete Fourier transforms, respectively, $\odot$ represents element-wise multiplication, and $\mathcal{F}^*(\mathbf{M})$ is the complex conjugate of the Fourier-transformed microstructure.
This formulation enables a highly efficient computation of $S_2$, reducing the complexity from direct spatial averaging to a fast Fourier transform (FFT)-based approach. 
As a result, it is particularly well-suited for large 3D microstructure data.

We extended the GitHub implementation from \cite{cciccek_3d_2016} by integrating modules for efficient descriptor computation using accelerated PyTorch routines. 
Specifically, we leveraged the PyTorch neural network framework to compute the two-point correlation function $S_2$ directly on the GPU, enabling fast and scalable processing of high-resolution or large microstructures in a batched manner. 
For optimal performance during computation, we recommend wrapping the computation in a:
\begin{lstlisting}[language=Python]
with torch.no_grad():
    s2 = compute_s2(M)
\end{lstlisting} 
block to prevent unnecessary gradient tracking and memory overhead, especially when dealing with large volumes.

It is worth noting that gradient descriptor computation is primarily relevant during microstructure reconstruction, which is not supported within the published repository. 
For GPU-based reconstruction workflows, we instead refer to established libraries such as PyMKS~\cite{brough_materials_2017} and MCRpy~\cite{seibert_microstructure_2022-1}, which provide dedicated support for descriptor-based inverse modeling.

In this study, we derive average layer thicknesses using a probabilistic approach by analyzing interpolated $S_2$ spectra along the $z$-axis (stacking direction). 
Characteristic peaks in the autocorrelation signal are identified, and under the assumption of a Gaussian distribution, the standard deviation is estimated from the peak width to quantify uncertainty.
Lastly, nesting factors are calculated and compared to findings in~\cite{dureth_determining_2020}.

\section{Results}
\label{sec:results}
\subsection{In-situ Computer Tomography Experiments}
\label{ssec:insitu_exp}

\begin{figure*}
    \centering
    \input{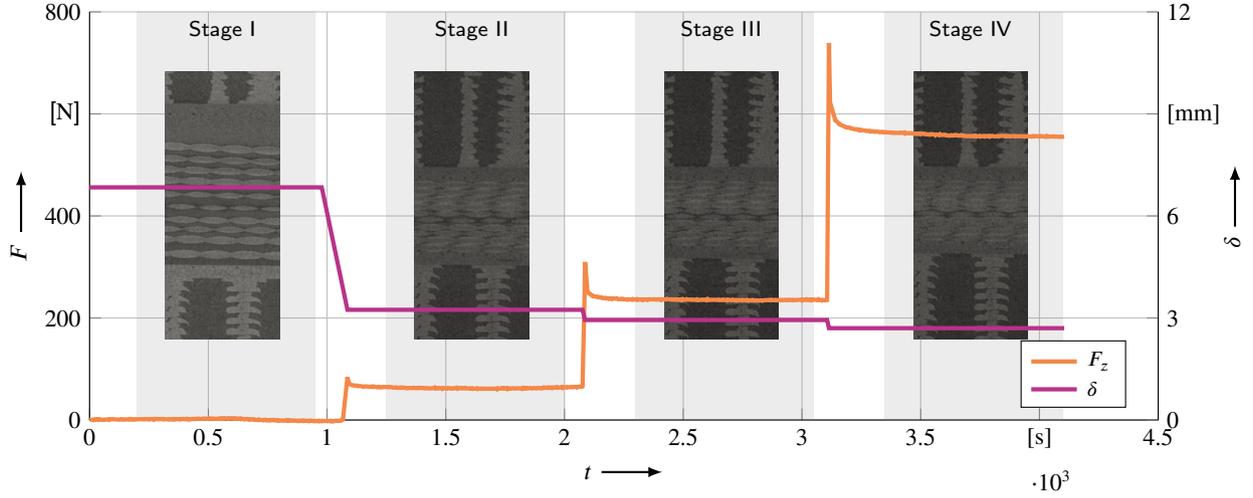}
    \caption{Illustration of compactation force $F$ and tamp gap $\delta$ over the measuring time $t$ as well as center slice of the shape $[600,1,256]$ of each CT-stage with estimated FVC $\phi=$\qtylist[]{<30;50;55;60}{\percent} }
    \label{fig:t-F_delta_acq}
\end{figure*}

The in-situ CT experiments yielded reconstructions of sufficient quality to resolve key structural features, as illustrated in Fig.~\ref{fig:t-F_delta_acq}. 
Representative slices of size $[600,1,256]$ from the reconstructed volumes of the 10-layer stack are shown for each acquisition stage. 
The corresponding tamp gap $\delta$ and compaction force $F$ are plotted to indicate the mechanical loading conditions during imaging. 
For comparability with literature, the applied forces were translated into approximate pressures, yielding \qtylist{0; 24; 94; 222}{\kilo\pascal} for stages I through IV, respectively

However, compared to the reconstructed volumes from \cite{yousaf_compaction_2021} of textile reinforcements made from glass fiber at a slightly higher spatial resolution of \SI{18}{\micro\meter\per Voxel}, the contrast and noise level in our scans are significantly more challenging. Their dataset, acquired using 3120 projections, exhibits clearer yarn boundaries and improved signal-to-noise ratio, facilitating segmentation and quantitative analysis. 
In contrast, our scans, based on 1440 projections and a carbon fiber material with lower X-ray attenuation, present reduced contrast and increased reconstruction noise. 
Despite these limitations, key structural features remain identifiable across all compaction stages, enabling meaningful analysis using our 3D-UNet-based segmentation approach. 
The applied machine learning pipeline effectively compensates for lower contrast and increased noise, allowing robust extraction of relevant geometric features even under suboptimal imaging conditions.

\subsection{3D-UNet Accuracy Evaluation}
\label{ssec:3dunet_evaluation}

\begin{table*}[htb]
    \centering
    \caption{Results of evaluation metrics $F1$ and \texttt{meanIoU} for stages I to IV of a 10-layer stacked CT-scan; more details about the evaluation and training datasets in~\cite{dureth_torch3dseg_2025}}
    \label{tab:eval_metrics}
 \begin{tabular}{ll|ccc|ccc}
\toprule
\textbf{Stage} & \textbf{Dataset} & \multicolumn{3}{c|}{\textbf{MeanIoU}} & \multicolumn{3}{c}{\textbf{F1Score}} \\
 &  & Class 0 (matrix) & Class 1 (weft) & Class 2 (fill) & Class 0 (matrix) & Class 1 (weft) & Class 2 (fill) \\
\midrule
\multirow{2}{*}{Stage I} & train & \num{0.9941} &  \num{0.9232} &  \num{0.9352} &  \num{0.9971} &  \num{0.9601} &  \num{0.9665} \\
                         & eval  & \num{0.9940} &  \num{0.9227} &  \num{0.9346} &  \num{0.9970} &  \num{0.9598} &  \num{0.9662} \\
\midrule
\multirow{2}{*}{Stage II} & train & \num{0.9924} &  \num{0.9063} &  \num{0.9086} &  \num{0.9962} &  \num{0.9509} &  \num{0.9521} \\
                          & eval  & \num{0.9839} &  \num{0.8565} &  \num{0.8480} &  \num{0.9919} &  \num{0.9227} &  \num{0.9177} \\
\midrule
\multirow{2}{*}{Stage III} & train & \num{0.8988} &  \num{0.8784} &  \num{0.8922} &  \num{0.9467} &  \num{0.9353} &  \num{0.9430} \\
                           & eval  & \num{0.8976} &  \num{0.8092} &  \num{0.8188} &  \num{0.9460} &  \num{0.8946} &  \num{0.9004} \\
\midrule
\multirow{2}{*}{Stage IV} & train & \num{0.9349} &  \num{0.8670} &  \num{0.8707} &  \num{0.9664} &  \num{0.9288} &  \num{0.9309} \\
                          & eval  & \num{0.9372} &  \num{0.8217} &  \num{0.8286} &  \num{0.9676} &  \num{0.9021} &  \num{0.9063} \\
\bottomrule
\end{tabular}
\end{table*}

The training process exhibited rapid convergence of the loss function within the 50-epoch training schedule. 
The model was trained over a total duration of \SI{3.4}{\hour} using two GPUs, which enabled efficient experimentation and facilitated timely parameter tuning throughout the development phase. 
As elaborated in Section~\ref{ssec:model_unet}, the adjustments in the model architecture proved advantageous, contributing to effective loss minimization. 
The final training and evaluation losses, computed using Eq.~\ref{eq:loss}, reached \num{0.5335} and \num{0.6492}, respectively.

After training, the model achieved an $F1$ score of 0.9430 and a mean IoU of 0.5704 on the training set. 
On the evaluation set, these metrics dropped to \num{0.5650} and \num{0.4249}, respectively, indicating limited generalization performance under evaluation conditions, apparently. 
Nevertheless, in full-volume prediction mode, the model exhibits significantly improved performance. 
A sliding-window inference strategy with a stride of 48 voxels is employed to process the complete dataset. 
This approach leads to overlapping of 80 voxels and \SI{62.5}{\percent}, respectively, per prediction patch, which are subsequently averaged across windows. 
The resulting aggregation reduces local inconsistencies and improves segmentation accuracy, particularly in challenging regions and near boundaries.
This improvement was already observed in medical segmentation tasks as shown in~\cite{kamnitsas_efficient_2017,isensee_nnu-net_2021}, where a gradually decreasing accuracy towards patch edges could be observed. 
To mitigate this, a sliding-window inference strategy with overlap, like presented in this work, and a Gaussian-weighted averaging scheme were employed to produce the final prediction, significantly enhancing consistency and boundary accuracy.
Additionally, the use of a mirror padding of 16 voxels at the volume boundaries helps to reduce edge artifacts and significantly improves segmentation accuracy near volume margins as shown in~\cite{cciccek_3d_2016}.
These supporting mechanisms are not implemented in the pure training pipeline, which results in lower accuracy and higher false-positive and false-negative rates at the boundaries.
\begin{figure*}
    \centering
    \begin{minipage}[t]{0.7\textwidth}
        \centering
        \subfloat[]{\begin{tikzpicture}[trim axis left, trim axis right]
    \begin{axis}[
        axis on top,
        enlargelimits=false,
        xmin=0,xmax=256,
        xlabel={$x$ \tikzArrow}, ylabel={$z$ \tikzArrow},
        xtick={0,64,128,192,256},
        xticklabels={$0$,$64$,$128$,$[\si{\Px}]$,$256$},
        ytick={0,64,128,192,256},
        yticklabels={$0$,$64$,$128$,$[\si{\Px}]$,$256$},
        width=.4\textwidth,
        height=.4\textwidth,
        tick style={white},  
        colormap={plasma}{ 
            rgb=(0.050383,0.029803,0.527975)
            rgb=(0.213753,0.000000,0.800000)
            rgb=(0.497000,0.000000,0.831373)
            rgb=(0.735683,0.000000,0.745189)
            rgb=(0.901961,0.294118,0.294118)
            rgb=(0.986000,0.730000,0.184000)
            rgb=(0.940015,0.975158,0.131326)
        },
        colorbar, 
        colorbar style={
            width=0.3cm,  
            ylabel={$p(c=1)$},  
            ylabel style={at={(0.1,.5)},anchor=south,font=\footnotesize,text=black},
            ytick={0, 0.25, 0.5, 0.75, 1.0},
            yticklabels={$0$, $0.25$, $0.5$, $[-]$, $1.0$},yticklabel style={font=\footnotesize,text=black}},
    ]
    
    \addplot graphics [xmin=0, xmax=256, ymin=0, ymax=256] {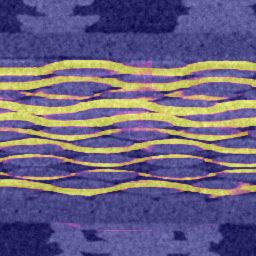};
    
    \end{axis}
\end{tikzpicture}\label{fig:eval_prediction_results:a}}\hspace{8em}
        \subfloat[]{\begin{tikzpicture}[trim axis left, trim axis right]
    \begin{axis}[
        axis on top,
        enlargelimits=false,
        xlabel={$x$ \tikzArrow}, ylabel={$z$ \tikzArrow},
        xtick={0,64,128,192,256},
        xticklabels={$0$,$64$,$128$,$[\si{\Px}]$,$256$},
        ytick={0,64,128,192,256},
        yticklabels={$0$,$64$,$128$,$[\si{\Px}]$,$256$},
        width=.4\textwidth,
        height=.4\textwidth,
        tick style={white},  
        colormap={plasma}{ 
            rgb=(0.050383,0.029803,0.527975)
            rgb=(0.213753,0.000000,0.800000)
            rgb=(0.497000,0.000000,0.831373)
            rgb=(0.735683,0.000000,0.745189)
            rgb=(0.901961,0.294118,0.294118)
            rgb=(0.986000,0.730000,0.184000)
            rgb=(0.940015,0.975158,0.131326)
        },
        colorbar, 
         colorbar style={
            width=0.3cm,  
            ylabel={$\Delta p(c=1)$},  
            ylabel style={at={(0.1,.5)},anchor=south,font=\footnotesize,text=black},
            ytick={0, 0.25, 0.5, 0.75, 1.0},
            yticklabels={$-0.8$, $-0.4$, $0.0$, $[-]$, $0.8$},yticklabel style={font=\footnotesize,text=black}},
    ]
           \addplot graphics [xmin=0, xmax=256, ymin=0, ymax=256] {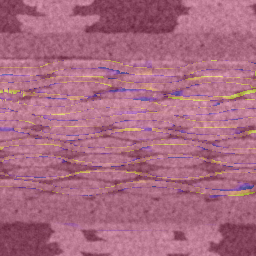};
    \end{axis}
\end{tikzpicture}\label{fig:eval_prediction_results:b}}\\[2ex]
        \subfloat[]{\begin{tikzpicture}[trim axis left, trim axis right]
    \begin{axis}[
        axis on top,
        enlargelimits=false,
        xlabel={$x$ \tikzArrow}, ylabel={$z$ \tikzArrow},
        xtick={0,64,128,192,256},
        xticklabels={$0$,$64$,$128$,$[\si{\Px}]$,$256$},
        ytick={0,64,128,192,256},
        yticklabels={$0$,$64$,$128$,$[\si{\Px}]$,$256$},
        width=.4\textwidth,
        height=.4\textwidth,
        tick style={white},  
        colormap={plasma}{ 
            rgb=(0.050383,0.029803,0.527975)
            rgb=(0.213753,0.000000,0.800000)
            rgb=(0.497000,0.000000,0.831373)
            rgb=(0.735683,0.000000,0.745189)
            rgb=(0.901961,0.294118,0.294118)
            rgb=(0.986000,0.730000,0.184000)
            rgb=(0.940015,0.975158,0.131326)
        },
        colorbar, 
        colorbar style={
            width=0.3cm,  
            ylabel={$p(c=2)$},  
            ylabel style={at={(0.1,.5)},anchor=south,font=\footnotesize,text=black},
            ytick={0, 0.25, 0.5, 0.75, 1.0},
            yticklabels={$0$, $0.25$, $0.5$, $[-]$, $1.0$},yticklabel style={font=\footnotesize,text=black}},
    ]
           \addplot graphics [xmin=0, xmax=256, ymin=0, ymax=256] {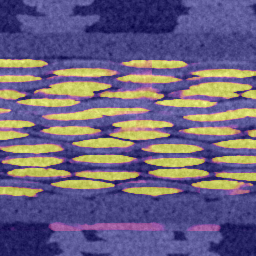};
    \end{axis}
\end{tikzpicture}\label{fig:eval_prediction_results:c}}\hspace{8em}
        \subfloat[]{\begin{tikzpicture}[trim axis left, trim axis right]
    \begin{axis}[
        axis on top,
        enlargelimits=false,
        xlabel={$x$ \tikzArrow}, ylabel={$z$ \tikzArrow},
        xtick={0,64,128,192,256},
        xticklabels={$0$,$64$,$128$,$[\si{\Px}]$,$256$},
        ytick={0,64,128,192,256},
        yticklabels={$0$,$64$,$128$,$[\si{\Px}]$,$256$},
        width=.4\textwidth,
        height=.4\textwidth,
        tick style={white},  
        colormap={plasma}{ 
            rgb=(0.050383,0.029803,0.527975)
            rgb=(0.213753,0.000000,0.800000)
            rgb=(0.497000,0.000000,0.831373)
            rgb=(0.735683,0.000000,0.745189)
            rgb=(0.901961,0.294118,0.294118)
            rgb=(0.986000,0.730000,0.184000)
            rgb=(0.940015,0.975158,0.131326)
        },
        colorbar, 
        colorbar style={
            width=0.3cm,  
            ylabel={$\Delta p(c=2)$},  
            ylabel style={at={(0.1,.5)},anchor=south,font=\footnotesize,text=black},
            ytick={0, 0.25, 0.5, 0.75, 1.0},
            yticklabels={$-0.8$, $-0.4$, $0.0$, $[-]$, $0.8$},yticklabel style={font=\footnotesize,text=black}},
    ]
           \addplot graphics [xmin=0, xmax=256, ymin=0, ymax=256] {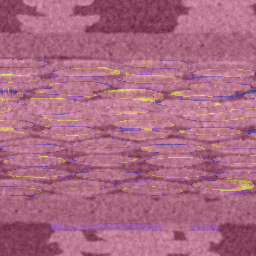};
    \end{axis}
\end{tikzpicture}\label{fig:eval_prediction_results:d}}
    \end{minipage}%
    \hfill
    \begin{minipage}[t]{0.3\textwidth}
        \vspace*{4em}  
        \centering
        \subfloat[]{\includegraphics[width=\textwidth]{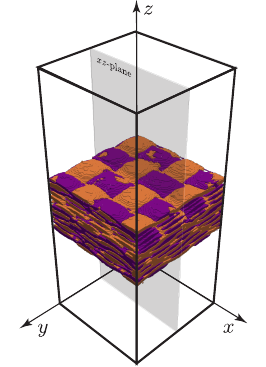}\label{fig:eval_prediction_results:e}}
    \end{minipage}

    \caption{Prediction results on a 10-layer evaluation dataset (stage IV) at the center $xz$-slice (cropped in $z$ for clarity): \protect\subref{fig:eval_prediction_results:a},\protect\subref{fig:eval_prediction_results:c} show predicted probabilities $p(c)$ for classes 1 and 2; \protect\subref{fig:eval_prediction_results:b},\protect\subref{fig:eval_prediction_results:d} show differences to the ground truth probability $\Delta p(c)$ for each class; all results are overlayed to the raw CT-data ;\protect\subref{fig:eval_prediction_results:e} displays the segmented volume with the highlighted $xz$-slice}
    \label{fig:eval_prediction_results}
\end{figure*}
Table~\ref{tab:eval_metrics} summarizes the $F1$ score and mean IoU achieved for each class across all compactation Stages I - IV, reported separately for the training and evaluation datasets. 
Overall, the model performs best on class 0 (matrix/background), while classes 1 (weft) and 2 (fill) show comparable but slightly lower accuracy.
A clear decline in performance is observed with increasing compaction from Stage I to  IV. 
Specifically, the mean IoU for classes 1 and 2 on the evaluation set decreases from \num{0.9227} and \num{0.9346} to \num{0.8217} and \num{0.8286}, respectively.
For the $F1$ score, a similar decrease is observable. 
Here the metric drops from \num{.9598} and \num{0.9662} to \num{0.9021} and \num{0.9063}
 for classes 1 and 2, respectively.

Figure~\ref{fig:eval_prediction_results} compares the prediction results to the ground truth labels for the center slice in the $xz$-plane of a 10-layer scan from the evaluation dataset at stage IV.
Here, the obtained class probabilities $p(c)$ for classes 1 and 2 are shown in subfigures~\subref{fig:eval_prediction_results:a} and~\subref{fig:eval_prediction_results:c}, respectively. 
The corresponding differences between predicted and ground truth probabilities highlight local deviations and are visualized in subfigures~\subref{fig:eval_prediction_results:b} and~\subref{fig:eval_prediction_results:d}. 
Subfigure~\subref{fig:eval_prediction_results:e} provides spatial context by depicting the segmented 3D volume with the center $xz$-slice highlighted.
While the predictions generally align well with the ground truth and demonstrate high spatial accuracy, elevated differences can be observed, particularly at class boundaries and along the edges of segmented yarns, indicating localized uncertainty in the transition regions.
However, it is important to note that manual labeling itself introduces uncertainty, especially in regions with low contrast or diffuse transitions, which may partly account for the observed discrepancies and limit the precision of the ground truth reference.

\subsection{Descriptor-based Analysis}
\label{ssec:descriptor_analysis}

\begin{figure*}[htb]
    \centering
    \hspace{-3em}
    \subfloat[]{\raisebox{.25\height}{\includegraphics[page=1,width=.15\textwidth]{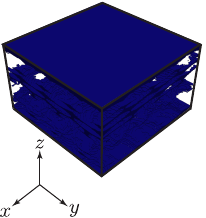}}\label{fig:D_s2_285:a}}\hspace{3em}
    \subfloat[]{\begin{tikzpicture}[trim axis left, trim axis right]
    \begin{axis}[
        axis on top,
        enlargelimits=false,
        xlabel={$x$ \tikzArrow}, ylabel={$y$ \tikzArrow},
        ylabel style={yshift=-6pt},
        xtick={0, 37, 75, 112, 149},
        xticklabels={$-75$, $-38$, $0$, $[{\si{\Px}}]$, $75$},
        ytick={0, 37, 75, 112, 149},
        yticklabels={$-75$, $-38$, $0$, $[{\si{\Px}}]$, $75$},
        width=.25\textwidth,
        height=.25\textwidth,
        tick style={white},  
    ]
        \addplot graphics [xmin=0, xmax=150, ymin=0, ymax=150] {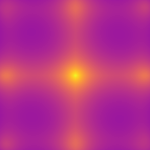};
    \end{axis}
\end{tikzpicture}\label{fig:D_s2_285:b}}\hspace{3em}
    \subfloat[]{\begin{tikzpicture}[trim axis left, trim axis right]
    \begin{axis}[
        colormap={plasma}{ 
            rgb=(0.050383,0.029803,0.527975)
            rgb=(0.213753,0.000000,0.800000)
            rgb=(0.497000,0.000000,0.831373)
            rgb=(0.735683,0.000000,0.745189)
            rgb=(0.901961,0.294118,0.294118)
            rgb=(0.986000,0.730000,0.184000)
            rgb=(0.940015,0.975158,0.131326)
        },
        axis on top,
        enlargelimits=false,
        xlabel={$x$ \tikzArrow}, ylabel={$z$ \tikzArrow},
        ylabel style={yshift=-6pt},
        xtick={0, 37, 75, 112, 149},
        xticklabels={$-75$, $-38$, $0$, $[{\si{\Px}}]$, $75$},
        ytick={0, 37, 75, 112, 149},
        yticklabels={$-75$, $-38$, $0$, $[{\si{\Px}}]$, $75$},
        width=.25\textwidth,
        height=.25\textwidth,
        tick style={white},  
    ]
\addplot graphics [xmin=0, xmax=150, ymin=0, ymax=150] {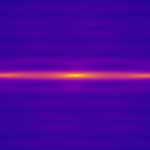};
\draw[white,dashed, thick] (axis cs:75,0) -- (axis cs:75,150);
    \end{axis}
\end{tikzpicture}\label{fig:D_s2_285:c}}\hspace{3em}
    \subfloat[]{\begin{tikzpicture}[trim axis left, trim axis right]
    \begin{axis}[
        colormap={plasma}{ 
            rgb=(0.050383,0.029803,0.527975)
            rgb=(0.213753,0.000000,0.800000)
            rgb=(0.497000,0.000000,0.831373)
            rgb=(0.735683,0.000000,0.745189)
            rgb=(0.901961,0.294118,0.294118)
            rgb=(0.986000,0.730000,0.184000)
            rgb=(0.940015,0.975158,0.131326)
        },
        axis on top,
        enlargelimits=false,
        colorbar, 
        colorbar style={
            width=0.3cm,  
            ylabel={$S_2/\max\{S_2\}$},  
            ylabel style={at={(0.1,.5)},anchor=south,font=\footnotesize,text=black},
            ytick={0, 0.25, 0.5, 0.75, 1.0},
            yticklabels={$0$, $0.25$, $0.5$, $[-]$, $1.0$},yticklabel style={font=\footnotesize},text=black},
        xlabel={$y$ \tikzArrow}, ylabel={$z$ \tikzArrow},
        ylabel style={yshift=-6pt},
        xtick={0, 37, 75, 112, 149},
        xticklabels={$-75$, $-38$, $0$, $[{\si{\Px}}]$, $75$},
        ytick={0, 37, 75, 112, 149},
        yticklabels={$-75$, $-38$, $0$, $[{\si{\Px}}]$, $75$},
        width=.25\textwidth,
        height=.25\textwidth,
        tick style={white},  
    ]
\addplot graphics [xmin=0, xmax=150, ymin=0, ymax=150] {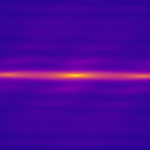};
\draw[white,dashed, thick] (axis cs:75,0) -- (axis cs:75,150);
    \end{axis}
\end{tikzpicture}\label{fig:D_s2_285:d}}\\
    \hspace{-3em}
    \subfloat[]{\raisebox{.25\height}{\includegraphics[page=3,width=.15\textwidth]{figures/fig_D_s2_160/285_10-layer_03_00_phases.pdf}}\label{fig:D_s2_285:e}}\hspace{3em}
    \subfloat[]{\begin{tikzpicture}[trim axis left, trim axis right]
    \begin{axis}[
        colormap={plasma}{ 
            rgb=(0.050383,0.029803,0.527975)
            rgb=(0.213753,0.000000,0.800000)
            rgb=(0.497000,0.000000,0.831373)
            rgb=(0.735683,0.000000,0.745189)
            rgb=(0.901961,0.294118,0.294118)
            rgb=(0.986000,0.730000,0.184000)
            rgb=(0.940015,0.975158,0.131326)
        },
        axis on top,
        enlargelimits=false,
        xlabel={$x$ \tikzArrow}, ylabel={$y$ \tikzArrow},
        ylabel style={yshift=-6pt},
        xtick={0, 37, 75, 112, 149},
        xticklabels={$-75$, $-38$, $0$, $[{\si{\Px}}]$, $75$},
        ytick={0, 37, 75, 112, 149},
        yticklabels={$-75$, $-38$, $0$, $[{\si{\Px}}]$, $75$},
        width=.25\textwidth,
        height=.25\textwidth,
        tick style={white},  
    ]
        \addplot graphics [xmin=0, xmax=150, ymin=0, ymax=150] {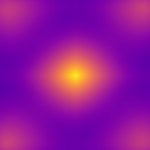};
    \end{axis}
\end{tikzpicture}\label{fig:D_s2_285:f}}\hspace{3em}
    \subfloat[]{\begin{tikzpicture}[trim axis left, trim axis right]
    \begin{axis}[
        colormap={plasma}{ 
            rgb=(0.050383,0.029803,0.527975)
            rgb=(0.213753,0.000000,0.800000)
            rgb=(0.497000,0.000000,0.831373)
            rgb=(0.735683,0.000000,0.745189)
            rgb=(0.901961,0.294118,0.294118)
            rgb=(0.986000,0.730000,0.184000)
            rgb=(0.940015,0.975158,0.131326)
        },
        axis on top,
        enlargelimits=false,
        xlabel={$x$ \tikzArrow}, ylabel={$z$ \tikzArrow},
        ylabel style={yshift=-6pt},
        xtick={0, 37, 75, 112, 149},
        xticklabels={$-75$, $-38$, $0$, $[{\si{\Px}}]$, $75$},
        ytick={0, 37, 75, 112, 149},
        yticklabels={$-75$, $-38$, $0$, $[{\si{\Px}}]$, $75$},
        width=.25\textwidth,
        height=.25\textwidth,
        tick style={white},  
    ]
\addplot graphics [xmin=0, xmax=150, ymin=0, ymax=150] {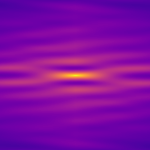};
\draw[white,dashed, thick] (axis cs:75,0) -- (axis cs:75,150);
    \end{axis}
\end{tikzpicture}\label{fig:D_s2_285:g}}\hspace{3em}
    \subfloat[]{\begin{tikzpicture}[trim axis left, trim axis right]
    \begin{axis}[
               colormap={plasma}{ 
            rgb=(0.050383,0.029803,0.527975)
            rgb=(0.213753,0.000000,0.800000)
            rgb=(0.497000,0.000000,0.831373)
            rgb=(0.735683,0.000000,0.745189)
            rgb=(0.901961,0.294118,0.294118)
            rgb=(0.986000,0.730000,0.184000)
            rgb=(0.940015,0.975158,0.131326)
        },
        axis on top,
        enlargelimits=false,
        colorbar, 
        colorbar style={
            width=0.3cm,  
            ylabel={$S_2/\max\{S_2\}$},  
            ylabel style={at={(0.1,.5)},anchor=south,font=\footnotesize,text=black},
            ytick={0, 0.25, 0.5, 0.75, 1.0},
            yticklabels={$0$, $0.25$, $0.5$, $[-]$, $1.0$},yticklabel style={font=\footnotesize},text=black},
        xlabel={$y$ \tikzArrow}, ylabel={$z$ \tikzArrow},
        ylabel style={yshift=-6pt},
        xtick={0, 37, 75, 112, 149},
        xticklabels={$-75$, $-38$, $0$, $[{\si{\Px}}]$, $75$},
        ytick={0, 37, 75, 112, 149},
        yticklabels={$-75$, $-38$, $0$, $[{\si{\Px}}]$, $75$},
        width=.25\textwidth,
        height=.25\textwidth,
        tick style={white},  
    ]
\addplot graphics [xmin=0, xmax=150, ymin=0, ymax=150] {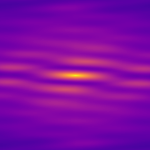};
\draw[white,dashed, thick] (axis cs:75,0) -- (axis cs:75,150);
    \end{axis}
\end{tikzpicture}\label{fig:D_s2_285:h}}\\
    \hspace{-3em}
    \subfloat[]{\raisebox{.25\height}{\includegraphics[page=2,width=.15\textwidth]{figures/fig_D_s2_160/285_10-layer_03_00_phases.pdf}}\label{fig:D_s2_285:i}}\hspace{3em}
    \subfloat[]{\begin{tikzpicture}[trim axis left, trim axis right]
    \begin{axis}[
        colormap={plasma}{ 
            rgb=(0.050383,0.029803,0.527975)
            rgb=(0.213753,0.000000,0.800000)
            rgb=(0.497000,0.000000,0.831373)
            rgb=(0.735683,0.000000,0.745189)
            rgb=(0.901961,0.294118,0.294118)
            rgb=(0.986000,0.730000,0.184000)
            rgb=(0.940015,0.975158,0.131326)
        },
        axis on top,
        enlargelimits=false,
        xlabel={$x$ \tikzArrow}, ylabel={$y$ \tikzArrow},
        ylabel style={yshift=-6pt},
        xtick={0, 37, 75, 112, 149},
        xticklabels={$-75$, $-38$, $0$, $[{\si{\Px}}]$, $75$},
        ytick={0, 37, 75, 112, 149},
        yticklabels={$-75$, $-38$, $0$, $[{\si{\Px}}]$, $75$},
        width=.25\textwidth,
        height=.25\textwidth,
        tick style={white},  
    ]
        \addplot graphics [xmin=0, xmax=150, ymin=0, ymax=150] {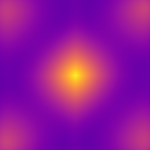};
    \end{axis}
\end{tikzpicture}\label{fig:D_s2_285:j}}\hspace{3em}
    \subfloat[]{\begin{tikzpicture}[trim axis left, trim axis right]
    \begin{axis}[
        colormap={plasma}{ 
            rgb=(0.050383,0.029803,0.527975)
            rgb=(0.213753,0.000000,0.800000)
            rgb=(0.497000,0.000000,0.831373)
            rgb=(0.735683,0.000000,0.745189)
            rgb=(0.901961,0.294118,0.294118)
            rgb=(0.986000,0.730000,0.184000)
            rgb=(0.940015,0.975158,0.131326)
        },
        axis on top,
        enlargelimits=false,
        xlabel={$x$ \tikzArrow}, ylabel={$z$ \tikzArrow},
        ylabel style={yshift=-6pt},
        xtick={0, 37, 75, 112, 149},
        xticklabels={$-75$, $-38$, $0$, $[{\si{\Px}}]$, $75$},
        ytick={0, 37, 75, 112, 149},
        yticklabels={$-75$, $-38$, $0$, $[{\si{\Px}}]$, $75$},
        width=.25\textwidth,
        height=.25\textwidth,
        tick style={white},  
    ]
\addplot graphics [xmin=0, xmax=150, ymin=0, ymax=150] {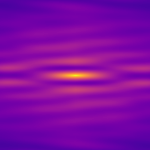};

\draw[white,dashed, thick] (axis cs:75,0) -- (axis cs:75,150);

    \end{axis}
\end{tikzpicture}\label{fig:D_s2_285:k}}\hspace{3em}
    \subfloat[]{\begin{tikzpicture}[trim axis left, trim axis right]
    \begin{axis}[
               colormap={plasma}{ 
            rgb=(0.050383,0.029803,0.527975)
            rgb=(0.213753,0.000000,0.800000)
            rgb=(0.497000,0.000000,0.831373)
            rgb=(0.735683,0.000000,0.745189)
            rgb=(0.901961,0.294118,0.294118)
            rgb=(0.986000,0.730000,0.184000)
            rgb=(0.940015,0.975158,0.131326)
        },
        axis on top,
        enlargelimits=false,
        colorbar, 
        colorbar style={
            width=0.3cm,  
            ylabel={$S_2/\max\{S_2\}$},  
            ylabel style={at={(0.1,.5)},anchor=south,font=\footnotesize,text=black},
            ytick={0, 0.25, 0.5, 0.75, 1.0},
            yticklabels={$0$, $0.25$, $0.5$, $[-]$, $1.0$},yticklabel style={font=\footnotesize},text=black},
        xlabel={$y$ \tikzArrow}, ylabel={$z$ \tikzArrow},
        ylabel style={yshift=-6pt},
        xtick={0, 37, 75, 112, 149},
        xticklabels={$-75$, $-38$, $0$, $[{\si{\Px}}]$, $75$},
        ytick={0, 37, 75, 112, 149},
        yticklabels={$-75$, $-38$, $0$, $[{\si{\Px}}]$, $75$},
        width=.25\textwidth,
        height=.25\textwidth,
        tick style={white},  
    ]
\addplot graphics [xmin=0, xmax=150, ymin=0, ymax=150] {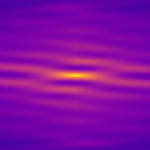};
\draw[white,dashed, thick] (axis cs:75,0) -- (axis cs:75,150);   
    \end{axis}
\end{tikzpicture}\label{fig:D_s2_285:l}}\\ 
    
    \caption{Illustration of normalized descriptor-based analysis via 2-point correlation function $S_2$, computed from the predicted segmentation (Figure~\ref{fig:eval_prediction_results}) of the cropped reinforcement region. \protect\subref{fig:D_s2_285:a},\protect\subref{fig:D_s2_285:e},\protect\subref{fig:D_s2_285:i}  show segmentation masks with axes for matrix, weft, and fill. \protect\subref{fig:D_s2_285:b},\protect\subref{fig:D_s2_285:f},\protect\subref{fig:D_s2_285:j}, (\protect\subref{fig:D_s2_285:c},\protect\subref{fig:D_s2_285:g},\protect\subref{fig:D_s2_285:k}), and \protect\subref{fig:D_s2_285:d},\protect\subref{fig:D_s2_285:h},\protect\subref{fig:D_s2_285:l} display center slices in the $xy$-, $xz$-, and $yz$-planes, respectively. 
    }
    \label{fig:D_s2_285}   
\end{figure*}

Figure~\ref{fig:D_s2_285} illustrates the 2-point correlation function $S_2$, as defined in Equation~\ref{eq:d_s2}, computed from the predicted segmentation volume shown in Figure~\ref{fig:eval_prediction_results}.
The volume is cropped along the z-axis to focus exclusively on the textile reinforcement region.
Subfigures~\subref{fig:D_s2_285:a},\subref{fig:D_s2_285:e}, and~\subref{fig:D_s2_285:i} display the corresponding segmentation masks and coordinate axes for the matrix, weft, and fill classes, respectively.
Subfigures~\subref{fig:D_s2_285:b},\subref{fig:D_s2_285:f}, and~\subref{fig:D_s2_285:j} show the center slices in the $xy$-plane, while the $xz$- and $yz$-planes are shown in Subfigures~\subref{fig:D_s2_285:c},\subref{fig:D_s2_285:g},\subref{fig:D_s2_285:k} and~\subref{fig:D_s2_285:d},\subref{fig:D_s2_285:h},\subref{fig:D_s2_285:l}, respectively.
All plots are limited to $\SI{\pm75}{voxel}$ or in spatial dimensions $\SI{\pm1.517}{\milli\meter}$ corresponding to \num{0.531} ondulations.

Although the segmentation achieves mean IoU of only \SI{82}{\percent}, the resulting spatial statistics remain highly consistent with the ground truth.
The mean squared error between the predicted and reference $S_2$ descriptors is just \num{2.19e-05}, indicating that the impact of voxel-level misclassifications on global microstructural correlation is negligible.

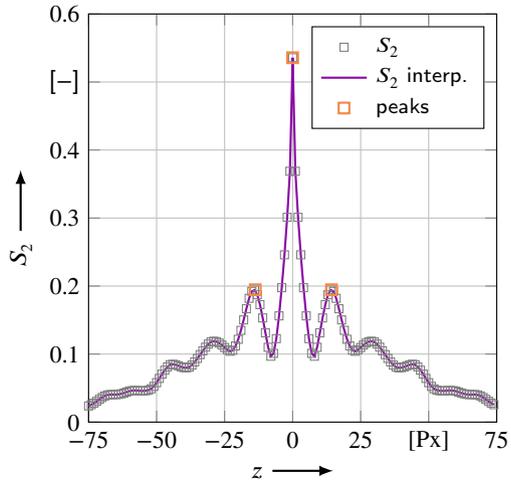
\begin{figure}[htb]
    \centering
    \begin{tikzpicture}[
                trim axis left,
                trim axis right,
                clip=True
                ]
    \begin{axis}[
                xmin=0, xmax=150,
                ymin=0, ymax=0.6,
                width=0.4*\textwidth, height=0.4*\textwidth,
             xtick={0,25,50,75,100,125,150}, 
            xticklabels={\num{-75},\num{-50},\num{-25},\num{0},\num{25},
                $[\si{\Px}]$,\num{75},},
                ytick={0,0.1,0.2,0.3,0.4,0.5,0.6}, 
                yticklabels={\num{0},\num{0.1},\num{0.2},\num{0.3},\num{0.4},[\si{-}],\num{0.6}},
                xlabel={$z$ \tikzArrow}, 
                ylabel={$S_2$ \tikzArrow},
                xmajorgrids,
                ymajorgrids,
                clip=false,
                legend pos=north east,
                legend cell align=left,
                legend style={cells={align=left}, font=\footnotesize}
                ]

\addplot[
    only marks,
    mark=square,
    mark size=1.5pt,
    color=gray,
] table[
    col sep=comma,
    x=x_original,
    y=spectrum
] {figures/285_10-layer_03_0.original.csv};
\addlegendentry{$S_2$}

    \addplot[
        thick,
        color=plasma4,
    ] table[
        col sep=comma,
        x=x_original,
        y=spectrum
    ] {figures/285_10-layer_03_0.original.csv};
    \addlegendentry{$S_2$ interp.}

\addplot[plasma7,     only marks,
    mark=square,thick] coordinates {(75, 0.5357) (89.2735,0.1946) (61.2768,0.1946)};
\addlegendentry{peaks}




    \end{axis}

\end{tikzpicture}
    \caption{Illustrative analysis of the $S_2$ spectrum along the $z$-axis, as shown in Fig.~\ref{fig:D_s2_285}, used to derive probabilistic estimates of the average layer thickness.}
    \label{fig:s2_analysis}
\end{figure}

A particularly insightful probabilistic evaluation from the $S_2$ descriptor involves analyzing the autocorrelation along the weft and fill directions in the $xz$- and $yz$-planes. 
By examining the spatial domain of the $S_2$ descriptor along the $z$-axis, assuming alignment with the textile structure, characteristic peaks corresponding to average layer thicknesses can be identified. 
Figure~\ref{fig:s2_analysis} shows the interpolated one-dimensional $S_2$ spectrum along the $z$-axis with identified peaks.

To improve peak resolution, the spectrum was interpolated using higher-order spline interpolation, and a Gaussian distribution was assumed around each peak to derive statistical estimates. 
In this analysis, the first dominant peak appeared at a $z$-directional distance of \SI{15.4159}{\Px} in the symmetric spectrum, corresponding to an average layer thickness of \SI{0.3092}{\milli\meter}. 
Under the Gaussian assumption, the peak width at \SI{99}{\percent} of the maximum was used to estimate the standard deviation, yielding a probabilistic layer thickness of \SI{15.2901(5869)}{\Px} or \SI{0.3092(119)}{\milli\meter}, respectively.

For a 10-layer stacking configuration and a measured $\delta_{10}(\phi=0.6)$ value of \SI{2.7}{\milli\meter} (cf. Table~\ref{tab:test_matrix}), this yields a nesting factor (Eq.\ref{eq:nesting}) of \num{0.874(33)}. 
This value shows strong agreement with the previously reported value of \num[separate-uncertainty=true]{0.868(37)} in~\cite{dureth_determining_2020} from analysis of micrographs of a consolidated composite with FVC of approx. \SI{60}{\percent}. 

\begin{figure}[htb]
    \centering
    \begin{tikzpicture}[
    trim axis left,
    trim axis right,
    clip=true
]
    \begin{axis}[
        ybar,
        bar width=10pt,
        enlarge x limits=0.4,
        ymin=0.75, ymax=1,
        width=0.45\textwidth, height=0.4\textwidth,
        xlabel={$\phi\, [-]$ \tikzArrow},
        ylabel={$\mathcal{N}_F$ \tikzArrow},
        xtick={0.50,.55,0.65}, 
        xticklabels={\num{0.50},\num{0.55},\num{0.60}},
        ytick={0.75,0.8,0.85,0.9,0.95,1.0}, 
        yticklabels={\num{0.75},\num{0.80},\num{0.85},\num{0.90},[\si{-}],\num{1.00}},
        xtick=data,
        xmajorgrids,
        ymajorgrids,
        legend pos=south west,
        legend cell align=left,
        legend style={cells={align=left}, font=\footnotesize}
    ]

    \addplot+[
        ybar,
        bar shift=-10pt,
        fill=plasma4,
        error bars/.cd,
        y dir=both, y explicit
    ]
    coordinates {
        (0.5, 0.9397) +- (0, 0.04023247)
        (0.55, 0.8870) +- (0, 0.038360192)
        (0.6, 0.8540) +- (0, 0.039676623)
    };
    \addlegendentry{5-layer}

    \addplot+[
        ybar,
        bar shift=0pt,
        fill=plasma6,
        error bars/.cd,
        y dir=both, y explicit
    ]
    coordinates {
        (0.5, 0.9455) +- (0, 0.030086349)
        (0.55, 0.9159) +- (0, 0.031670372)
        (0.6, 0.8742) +- (0, 0.033581063)
    };
    \addlegendentry{10-layer}

    \addplot+[
        ybar,
        bar shift=10pt,
        fill=plasma7,
        error bars/.cd,
        y dir=both, y explicit
    ]
    coordinates {
        (0.5, 0.9420) +- (0, 0.034347388)
        (0.55, 0.9096) +- (0, 0.03721187)
        (0.6, 0.8729) +- (0, 0.040855063)
    };
    \addlegendentry{37-layer}

    \addplot+[
        ybar,
        bar shift=20pt,
        fill=gray,
        error bars/.cd,
        y dir=both, y explicit
    ]
    coordinates {
        (0.6, 0.868) +- (0, 0.037)
    };
    \addlegendentry{Düreth et al.~\cite{dureth_determining_2020}}
    
\addlegendimage{legend image code/.code={
            \draw[thick, gray] (0cm,0.0em) -- (0.3cm,0.0em);
            \draw[thick, gray] (0cm,-0.2em) -- (0.0cm,0.2em);
            \draw[thick, gray] (0.3cm,-0.2em) -- (0.3cm,0.2em);
            }
            }
\addlegendentry{std. dev.}

    \end{axis}
\end{tikzpicture}
    \caption{Results of the descriptor-based nesting analysis for a 5-, 10-, and 37-layer configurations.}
    \label{fig:nesting_results}
\end{figure}
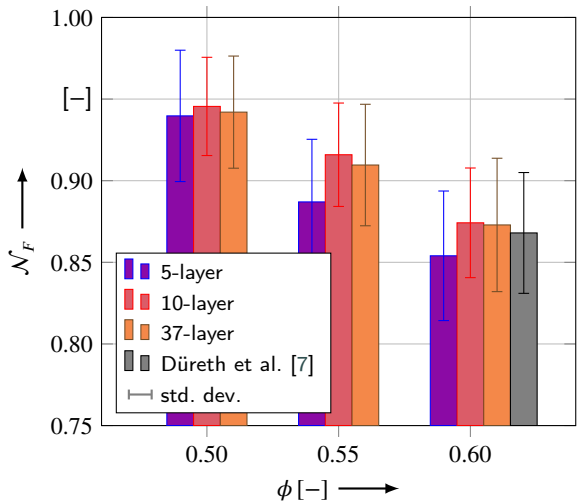

This brief study investigates the probabilistic nature of nesting in multilayer textile reinforcements, focusing on the potential existence of a nesting limit in 5-, 10-, and 37-layer configurations. 
The analysis considers several compaction stages within a fiber volume content (FVC) range of \qtyrange[]{50}{60}{\percent}. Figure~\ref{fig:nesting_results} presents the derived nesting factors in a statistical bar plot with corresponding uncertainty.
The results indicate that the 5-layer stack generally exhibits lower nesting factors, indicating a higher degree of interlayer nesting compared to configurations with a greater number of layers. 
While not a clear trend, there appears to be a tendency for increased stacking order to correlate with slightly lower nesting, particularly under higher compaction. 
However, the differences remain within the range of standard deviation, suggesting that although the 5-layer configuration shows more efficient nesting, higher stacking configurations may still exhibit comparable nesting behavior due to localized layer adaptation and increased interfacial contact.

\section{Discussion}
\label{sec:discussion}

The in-situ CT experiments provided volumetric data of adequate quality to support deep-learning-based segmentation across different compaction stages. 
Despite limitations in resolution and the presence of reconstruction noise, especially in higher compaction states, key structural features such as yarn trajectories remained visible, particularly in early stages with lower nesting complexity. 
However, as compaction increased, visual separation of individual layers became increasingly difficult due to overlapping yarns and reduced contrast. 
This underscores the need for continued refinement of imaging protocols in future studies, potentially including alternative cone-beam CT configurations or upgraded hardware setups tailored to the specific demands of textile composite analysis, in order to achieve higher spatial resolutions and improved structural clarity

Advanced reconstruction methods, such as those offered by the Tomographic Iterative GPU-based Reconstruction (TIGRE) Toolbox~\cite{biguri_tigre_2016}, present a promising direction. 
These methods have the potential to significantly enhance the visibility of fine structural features by increasing contrast and suppressing artifacts. 
In particular, they could improve segmentation quality in densely nested configurations like those observed in stage IV, where traditional reconstruction pipelines struggle to resolve internal geometries with sufficient clarity.

In addition to hardware and reconstruction advances, improving the reliability of training data remains critical. 
The segmentation models evaluated in this study were trained against manually labeled ground truth, which itself is subject to uncertainty, especially in transitional or low-contrast regions. 
As shown in Figure~\ref{fig:eval_prediction_results}, prediction differences tend to localize near class boundaries and fiber edges, where manual annotations may be ambiguous or inconsistent. 
This suggests that the apparent segmentation error includes not only model limitations but also uncertainty in the reference data.

To overcome this, synthetic data generation offers a viable alternative by enabling precise control over fiber architectures and guaranteeing perfect labels. 
Recent works such as those by Friemann et al.~\cite{friemann_automated_2024,friemann_synthetic_2025} and Vidal et al.~\cite{vidal_x-ray_2024} demonstrate the potential of simulation-based approaches to produce high-quality, annotated training volumes that reflect realistic textile structures. 
Integrating such data can improve model robustness and reduce dependency on labor-intensive manual annotation workflows.

Beyond voxel-wise classification, the UNet architecture used in this study (Fig.~\ref{fig:model_runet}) demonstrates the ability to learn broader geometric patterns -  high-dimensional features -  inherent to the textile architecture, including the repetitive, directional nature of yarn arrangements. 
This suggests that, with appropriate supervision, e.g., from synthetic data, such architectures could also be extended to approximate local orientation and anisotropy even in low-contrast CT scans. 
Currently, such detailed structural representations typically require high-resolution imaging with voxel sizes below \SIrange[]{2}{3}{\micro\meter} for gradient-based structure-tensor analysis~\cite{emerson_individual_2017,karamov_micro-ct_2020}. 
Leveraging deep learning models to infer this information from coarser, lower-resolution data would represent a significant advancement, enabling detailed structural analysis without the need for high-end imaging systems. 
This shift would further expand the role of segmentation networks from simple labeling tools to geometry-aware models capable of supporting downstream tasks such as yarn alignment quantification and anisotropic material modeling.

It is worth noting that hyperparameter tuning with tools like OmniOpt~\cite{winkler_omniopt_2021} and model architecture design remain challenging aspects of the segmentation pipeline. 
Given the model’s ability to learn high-dimensional geometric features, a reduction in the number of encoder and decoder blocks may be feasible. 
This could alleviate computational demands and allow for training on larger volumetric patches, thereby capturing broader spatial context more effectively.

Lastly, the analysis demonstrated that the average layer spacing and thickness could be identified reliably from the $S_2$ spectra in a probabilisitc manner, highlighting the method’s suitability for quantifying structural periodicity in woven composites. 
Beyond this, the $S_2$ descriptor holds potential for broader applications, such as comparing experimentally derived microstructures to idealized representative volume elements generated in tools like TexGen, or guiding microstructural reconstruction from statistical data, as already mentioned. 
Furthermore, the descriptor can be leveraged to characterize interlayer features that may influence failure mechanisms, including crack jumps or delamination pathways.

\section{Conclusion}
\label{sec:conclusion}

We presented a comprehensive 3D segmentation framework tailored for the structural analysis of low-resolution in-situ CT scans obtained during compaction experiments on dry carbon fiber textile reinforcements. 
The scans, acquired under experimental constraints at an approximate resolution of \SI{20}{voxel\per\micro\meter}, captured the progressive deformation of the textile architecture across multiple compaction stages.

By employing a modified 3D UNet architecture, the framework enabled robust voxel-wise segmentation of the volumetric data, successfully distinguishing key material constituents such as weft, fill, and matrix phases. 
Notably, the network demonstrated high generalization capabilities, accurately segmenting complex fiber arrangements even in high compaction conditions. 
In Stage IV, which featured a fiber volume content of approximately \SI{60}{\percent} and severe nesting, the model still achieved mean IoU scores of \num{0.8217} and \num{0.8286} for the weft and fill, respectively—despite reduced contrast and increased structural complexity. 
These results underscore the potential of geometry-aware deep learning models to support structural interpretation under realistic imaging constraints, reducing reliance on high-resolution systems.

Beyond segmentation, we introduced a descriptor-based analysis pipeline to quantitatively assess the textile architecture with a focus on nesting behavior.
Leveraging the two-point autocorrelation function and spectral analysis along the stacking direction, we derived probabilistic estimates of the average layer thickness and corresponding nesting factors. 
A nesting factor of \num{0.873(36)} was determined, which is in good agreement with the value of \num[separate-uncertainty=true]{0.868(37)} reported in~\cite{dureth_determining_2020}, thereby validating the effectiveness of the proposed approach.

The integration of deep learning segmentation and statistical descriptor analysis offers a powerful and non-destructive methodology for characterizing complex reinforcement structures in fiber composites. 
This approach not only enables quantitative assessment of key structural features, such as layer nesting, but also paves the way for future work on process-structure-property relationships, microstructure reconstruction, and model validation against idealized representative volume elements.
Ultimately, the methodology supports the design and optimization of textile-based composites in lightweight engineering applications under experimentally relevant conditions.

\appendix

\section{Appendix}

\begin{table}[ht]
\centering
\caption{3D-UNet architecture summary with tree structure; $-1$ denotes the batch size $b$; \num{9183245} parameters in total}
\label{tab:model_parameters}
\begin{tabular}{@{}>{\ttfamily}lcl@{}}
\toprule
\textbf{Layer (type: idx)} & \textbf{Output Shape} & \textbf{Param \#} \\
\midrule
3DUnet & $[-1, 3, 128, 128, 128]$ & -- \\
\midrule
Encoder Moduls & -- & -- \\
\hspace{1em}Encoder: 1 & $[-1, 48, 128, 128, 128]$ & 31,802 \\
\hspace{1em}Encoder: 2 & $[-1, 96, 64, 64, 64]$ & 187,200 \\
\hspace{1em}Encoder: 3 & $[-1, 192, 32, 32, 32]$ & 747,648 \\
\hspace{1em}Encoder: 4 & $[-1, 384, 16, 16, 16]$ & 2,988,288 \\
Decoder Moduls: & -- & -- \\
\hspace{1em}Decoder: 3 & $[-1, 192, 32, 32, 32]$ & 3,982,848 \\
\hspace{1em}Decoder: 2 & $[-1, 96, 64, 64, 64]$ & 996,096 \\
\hspace{1em}Decoder: 1 & $[-1, 48, 128, 128, 128]$ & 249,216 \\
Conv3d:  & $[-1, 3, 128, 128, 128]$ & 147 \\
Softmax: & $[-1, 3, 128, 128, 128]$ & -- \\
\bottomrule
\end{tabular}
\end{table}

\section*{Acknowledgements}
The research group of M. Gude gratefully acknowledges the support of the German Research Foundation (DFG). 
The experimental investigation, characterization, and descriptor-based analysis of textile reinforcements under compaction were funded through grant GU 614/30-1 (DFG project number: 450147819). 
The development of semantic segmentation methods for analyzing the material structure was supported by grant GU 614/35-1 (DFG project number: 496642725). 

The authors gratefully acknowledge the computing time made available to them on the high-performance computer at the NHR Center of TU Dresden. This center is jointly supported by the Federal Ministry of Education and Research and the state governments participating in the NHR (\url{www.nhr-verein.de/unsere-partner}).

The authors extend their sincere gratitude to Lars Klein\-kop for performing manual data annotation as a student assistant. 

\section*{Code Availability}
The code used for training, inference, evaluation, and descriptor-based analysis is publicly available at: \\
\url{https://github.com/choROPeNt/3dseg}.

This repository is a fork of the original PyTorch 3D-UNet implementation available at:  \\
\url{https://github.com/wolny/pytorch-3dunet}.

\printcredits

\bibliographystyle{elsarticle-num}

\bibliography{references}

\end{document}